\documentclass[conference]{IEEEtran}
\IEEEoverridecommandlockouts
\usepackage{cite}
\usepackage{amsmath,amssymb,amsfonts}
\usepackage{algorithmic}
\usepackage{graphicx}
\usepackage{textcomp}
\usepackage{xcolor}
\usepackage{comment}
\usepackage{gensymb}

\newcommand{\REV}[1]{{#1}}

\def\BibTeX{{\rm B\kern-.05em{\sc i\kern-.025em b}\kern-.08em
    T\kern-.1667em\lower.7ex\hbox{E}\kern-.125emX}}
\begin{document}

\title{Preprocessing Methods for Memristive Reservoir Computing for Image Recognition}

\author{\IEEEauthorblockN{Rishona Daniels\IEEEauthorrefmark{1}, Duna Wattad\IEEEauthorrefmark{1}, Ronny Ronen\IEEEauthorrefmark{1}, David Saad\IEEEauthorrefmark{4}, and Shahar Kvatinsky\IEEEauthorrefmark{1}}
\IEEEauthorblockA{\textit{\IEEEauthorrefmark{1}Viterbi Faculty of Electrical and Computer Engineering,} 
\textit{Technion - Israel Institute of Technology}, Haifa, Israel \\
\textit{\IEEEauthorrefmark{4}College of Engineering and Physical Sciences,} 
\textit{Aston University}, Birmingham B4 7ET, United Kingdom \\
rishonad@campus.technion.ac.il, duna.wattad@campus.technion.ac.il, ronny.ronen@technion.ac.il,\\ d.saad@aston.ac.uk,  shahar@ee.technion.ac.il}
}
\maketitle

\begin{abstract}
Reservoir computing (RC) has attracted attention as an efficient recurrent neural network architecture due to its simplified training, requiring only its last perceptron readout layer to be trained. When implemented with memristors, RC systems benefit from their dynamic properties, which make them ideal for reservoir construction. However, achieving high performance in memristor-based RC remains challenging, as it critically depends on the input preprocessing method and reservoir size. Despite growing interest, a comprehensive evaluation that quantifies the impact of these factors is still lacking. This paper systematically compares various preprocessing methods for memristive RC systems, assessing their effects on accuracy and energy consumption. We also propose a parity-based preprocessing method that improves accuracy by 2–6\% while requiring only a modest increase in device count compared to other methods. Our findings highlight the importance of informed preprocessing strategies to improve the efficiency and scalability of memristive RC systems.
\end{abstract}


\section{Introduction}

Neural networks have emerged as the leading solution for various computing tasks, including image recognition, natural language processing, and robotics. However, most traditional neural networks require time-consuming and power-intensive training. Reservoir computing (RC)~\cite{next_gen_rc}, a variant of recurrent neural networks (RNNs), addresses these challenges by training only the readout layer, thus simplifying training. RC has been effectively applied to model and forecast nonlinear dynamical systems, such as wind speed and direction forecasting \cite{wind_forecasting_using_rc}, and financial market prediction \cite{stock_market_prediction_using_rc}, while also proving useful in image \cite{img_recog_using_rc} and speech recognition~\cite{info_processing_in_single_dynamical_node}.

Compared to traditional RNNs such as long-short-term memory (LSTM), RC significantly reduces training computational overhead. This makes it useful for various classification and regression tasks, achieving state-of-the-art results in forecasting chaotic time series data, such as Mackey-Glass and Lorentz-63 \cite{chaotic_time_series_using_rc}. However, RC relies on complex reservoir dynamics, including \textit{echo-state properties} and \textit{fading memory}, which are computationally demanding when implemented using traditional digital computers. These implementations often require frequent memory updates and substantial data movement, leading to inefficiencies in energy and speed \cite{physical_rc_with_emerging_electronics_review}.

Inspired by the brain's processing capabilities, neuromorphic computing offers a promising alternative to standard digital designs by performing computations in the analog domain, directly within physical devices. This makes it an ideal platform for RC, enabling in-memory information processing directly in hardware more efficiently~\cite{physical_rc_with_emerging_electronics_review}. Neuromorphic hardware designs of RC have been demonstrated using various technologies, including analog circuits~\cite{delay_based_rc_using_analog_ckts}, FPGAs \cite{lsm_rc_using_fpga}, memristors \cite{rc_with_dynamic_memristors}, photonic devices \cite{photonic_rc_review}, and spintronic devices \cite{spintronic_rc}. Among these, memristors stand out because of their unique ability to combine storage and processing within the same physical location, alongside their small footprint.

Memristive RC has been successfully applied to tasks such as image classification \cite{rc_with_dynamic_memristors}, graph-based learning \cite{echo_state_graph_neural_network}, spiking neural networks \cite{snn_memristive_rc}, and time-series predictions \cite{rc_with_dynamic_memristors}. RC methods are typically divided into three main categories: Echo State Networks (ESN), Liquid State Machines (LSM) and Delay Feedback Network (DFN). The DFN approach, in particular, is popular for memristive RC systems due to its lower hardware complexity, reduced resource requirements, and improved speed \cite{rc_with_dynamic_memristors}, and, as such, is the focus of this paper. In the DFN method, the input data is fed into a reservoir of dynamical memristors that behave as the reservoir nodes -- individual processing units that collectively transform input signals into high-dimensional representations. However, accuracy depends heavily on the number of nodes and how the input of the reservoir nodes is preprocessed. Different preprocessing methods require a different number of reservoir nodes and result in varying accuracies. 

In this paper, we analyze and compare different preprocessing methods of DFN-based memristive RC systems, focusing on their impact on accuracy and energy consumption. We propose a parity-based preprocessing method in which we XOR the pixels of adjacent rows of the input image, resulting in higher accuracy of the reservoir computing network. The effectiveness of the parity approach is demonstrated using the MNIST dataset as a benchmark.

\section{Background}

\subsection{Reservoir Computing }
\begin{figure}[!t]
\centerline{\includegraphics[scale=0.29]{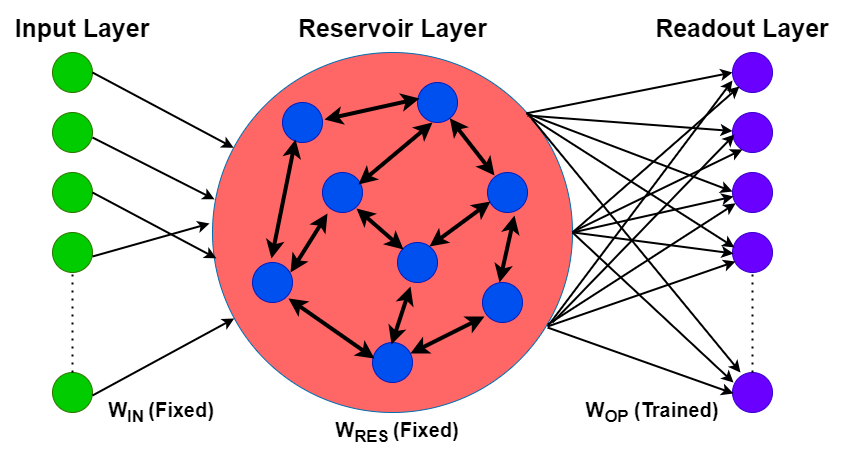}}
\caption{Block diagram of RC with input, reservoir, and readout layers. Only readout weights are trained; input and reservoir parameters are not trained.}
\label{fig_RC_Blk_diag}
\end{figure}
Reservoir computing (RC) is a specialized form of recurrent neural network. Recurrent networks exhibit complex non-linear dynamics owing to their coupling connections and the delays imposed by them \cite{rc_book_theory_physical_implementations_applications}. These dynamics enable the projection of input data into a high-dimension state space, which can be utilized to implement functions such as the classification of complex spatio-temporal input patterns. RC comprises three main blocks: the input, the reservoir, and the readout layer, as shown in Fig.~\ref{fig_RC_Blk_diag}.

In the input block, data such as images are converted into multiple spatio-temporal sequences through \textit{preprocessing}. This step effectively increases the feature dimensionality, helping to preserve important spatial characteristics of the original image. These enriched sequences are then fed into the reservoir nodes. The reservoir itself is a recurrent network exhibiting non-linear dynamics, often driven by oscillations or delays, similar to other neural network architectures. By projecting these spatio-temporal inputs into a high-dimensional dynamic state space, the reservoir enhances the system’s ability to recognize and process complex patterns.

For the reservoir to effectively represent the input data, it must exhibit two properties: the \textit{echo-state property} and the \textit{fading memory property}. Echo-state ensures that the reservoir retains the influence of past inputs over time, while the fading memory property prioritizes more recent inputs, giving them a greater impact on the state of the reservoir compared to older ones\cite{jaeger_echo_state_paper}. Finally, the readout layer (output layer) is a single perceptron that serves as a classifier and reads the reservoir's state. This layer is the only part of the system that is trained, typically using regression techniques. In contrast, the weights of the input and reservoir layers remain fixed throughout the supervised training process \cite{next_gen_rc}.


\subsection{Memristors}
A memristor is a two-terminal device that stores data as conductance~\cite{volatile_and_non_volatile_memristive_devices_for_neuromorphic_computing}. Its conductance can be adjusted using write voltage pulses to encode specific values and read by applying a pulse that does not alter the state. In nonvolatile memristors, the stored conductance remains stable until a new write pulse is applied. In contrast, volatile memristors gradually lose their conductance once the applied voltage drops below a threshold, known as the hold voltage (\textit{$V_{hold}$}). Thus, nonvolatile memristors suit long-term memory applications, while volatile ones are ideal for short-term memory. The retention time of a memristor refers to how long it retains its value. Nonvolatile memristors can store data for months to years, while volatile memristors have shorter retention times, ranging from nanoseconds to minutes.

In this paper, we employ the metal oxide volatile memristive model~\cite{model_of_Wox_memristor} whose I-V relationship is:
\begin{equation}
    I = (1-w)\alpha[1 - exp(-\beta V)] + w\gamma sinh(\delta V),
\label{IV}
\end{equation}
\begin{equation}
    \frac{dw}{dt} =\lambda sinh(\eta V) - \frac{w}{\tau},
\label{dw/dt}
\end{equation}
where $w \in [w_{Min}, w_{Max}]$ is the internal state variable of the memristor,
$\alpha$, $\beta$, $\gamma$, $\delta$, $\lambda$, and $\eta$ are positive fitting parameters dependent on material properties, and $\tau$ is the diffusion time constant that determines the rate of decay of $w$. 
When a positive write pulse is applied across the memristor (we define this case as $'1'$ pulse), Equation~(\ref{dw/dt}) can be approximated~\cite{rc_with_dynamic_memristors} to 
\begin{equation}
    \Delta w = R(w) \times t_{pulse} \times \lambda \times sinh(\eta \times V_{pulse}),
    \label{w_update}
\end{equation}
\begin{equation}
    R(w) = 1 - \frac{exp(3 \times w)}{exp(3\times w_{Max})},
    \label{R(w)}
\end{equation}
where $t_{pulse}$ is the pulse width, and $R(w)$ is the window function that constrains the value of $w$. Conversely, if no write pulse is applied across the memristor (we define this case as $'0'$ pulse), Equation~(\ref{dw/dt}) is approximated to reflect the volatile nature of the memristor such that $w$ decays: 
\begin{equation}
        \Delta w = (w - w_{Min}) \times \left(1 -  exp\left(-\frac{t_{pulse}}{\tau}\right)\right).
    \label{w_decay}
\end{equation}

\subsection{Reservoir Computing with Memristors}

Several approaches have been proposed for implementing reservoir computing with memristive devices. In \cite{novel_memristor_esn_MEM-ESN}, series and parallel combinations of connected memristors are employed to construct the reservoir. In \cite{hardware_esn_with_double_crossbar_memristors}, double crossbar arrays of randomly initialized nonvolatile memristors provide the recurrent dynamics of the reservoir. In \cite{memristor_based_LSM_with_insitu_training_method}, CMOS spiking leaky integrate-and-fire (LIF) neurons combined with memristive synapses constitute the reservoir (\textit{"liquid"}) layer. However, these methods tend to be hardware-intensive. Conversely, \cite{non_masking_based_rc_with_single_dynamic_memristor} uses a single volatile memristor fed by a histogram of oriented gradients (HOG) feature detector for image preprocessing. Although this approach reduces device count, it introduces complexity through the HOG preprocessing step and limits parallelism due to the single-device architecture. 

\begin{figure}[tb]
\centering
\includegraphics[scale=0.195]{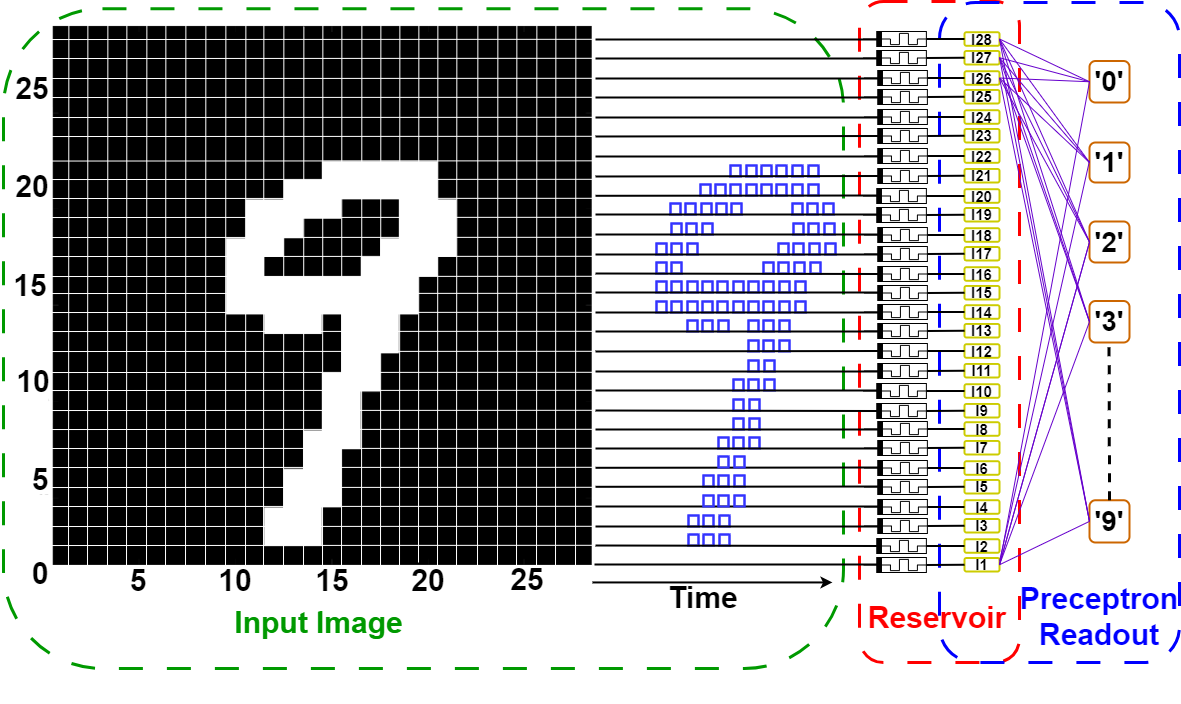}
\caption{Schematic of delay feedback network (DFN) reservoir computing (RC). Each row of the binary input image is converted to a pulse train. The pulse train is then temporally written into the volatile memristors that form the reservoir. After the entire image is written, a read pulse is applied to all volatile memristors, and the obtained currents are rescaled to form the activations of the perceptron readout layer.}
\label{fig_RC_DFN}
\end{figure}

A prominent method, the Delay Feedback Network (DFN, see Fig.~\ref{fig_RC_DFN}), addresses these trade-offs. It uses fewer devices than ESN and LSM methods described in \cite{novel_memristor_esn_MEM-ESN, hardware_esn_with_double_crossbar_memristors, memristor_based_LSM_with_insitu_training_method}, and enables parallel data input and processing with relatively simple preprocessing methods, unlike \cite{non_masking_based_rc_with_single_dynamic_memristor}.
In DFN, the reservoir uses volatile memristors with a slow decay rate. Preprocessing converts each image row into spatio-temporal write-pulse sequences, each written to a volatile memristor. As these input sequences are written, the memristors retain their value (echo state property) while decaying at a certain fixed rate (fading memory). The decay rate needs to be properly tuned to faithfully represent the input data.  After the entire input sequence is written, a read pulse retrieves the final memristor state, producing currents that represent the final reservoir states. These currents are then passed to a perceptron layer.

In the perceptron layer, the current values are multiplied by learnable weights, and activation (e.g., softmax) is applied to produce classification results. In training, a loss function compares these results to the true labels, and the resulting error is used to update the weights of the readout layer. Conversely, the reservoir’s internal states are not modified by any training algorithm and, therefore, are not learned parameters. The final state of the reservoir is determined by the input data and the fading memory and echo state properties of the volatile memristors due to the inherent device physics. Once a new image is preprocessed and applied as input sequences to the volatile memristors, their states change according to the aforementioned properties and eventually completely decays before the next dataset sample arrives. Hence, the state of the reservoir memristors must be read immediately after the input image has been applied to prevent a loss of state information due to the natural decay of the device conductance. This requirement applies during both training and testing.

Implementing a multi-layer perceptron (MLP) with nonvolatile memristors as weights requires flattening the input image. 
In comparison, the DFN approach is more effective than a fully connected MLP because it avoids flattening the input image by using spatial-temporal processing, thereby reducing both the number of required memristors and the effective area. Moreover, since only the readout layer requires training, the energy and time for training are substantially lower. For example, if the input image is of size 28$\times$28 and an MLP with two hidden layers of 20 neurons each and an output layer with 10 neurons will require $28^2\times 20+20 \times 20+20 \times 10=16,280$ trainable parameters and the same number of memristors to represent these weights. The same operation can be performed using RC by 3652 memristors out of which 332 do not need to be trained, as we will show further. 

\REV{Prior relevant work includes \cite{rc_with_dynamic_memristors}, where varying input rate leads to different final memristor states, and \cite{2mem_1cap_rc}, which employs a reservoir built using kernels comprising two non-volatile memristors and a capacitor producing two responses to the same input. The capacitor sets the decay rate but adds area and energy overhead. 
}

\begin{table}[!t]
\caption{Preprocessing methods and Reservoir Sizes (assume input image with $n$ rows, $m$ columns, and $k$ sections)}
\begin{center}
\begin{tabular}{|c|c|c|}
\hline
\textbf{Dimension} &  \textbf{Parity} & \textbf{No. of Volatile Memristors} \\
\hline
1D & No & $n\times k$ \\
\hline
1D & Yes & $[n+(n-1)]\times k$  \\
\hline
2D & No & $(n+m)\times k$ \\
\hline
2D & Yes  & $[(n+m)+(n-1)]\times k$ \\
\hline

\hline
\end{tabular}
\label{rc_size}
\end{center}
\end{table}

\section{Preprocessing Methods}

\begin{figure*}[!t]
\centerline{\includegraphics[scale=0.32]{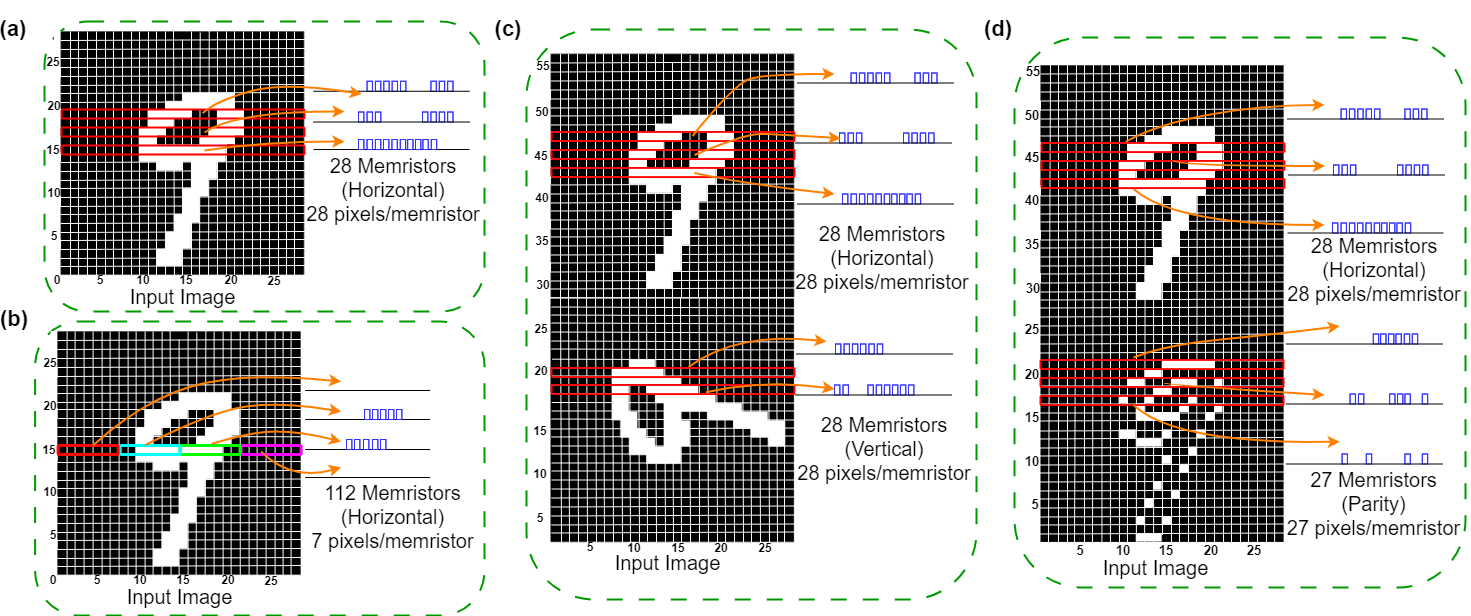}}
\caption{\REV{Schematics of preprocessing methods. (a) One-dimensional preprocessing: each horizontal row of pixels is converted into a pulse train of write voltages and is sequentially injected into a volatile memristor. (b) Input sectioning: only a portion of each row is given to each memristor. In the image shown, $4$ sections are used and seven pixels are applied to one memristor, improving accuracy by distributing the image representation across multiple devices. (c) Two-dimensional preprocessing: each horizontal row and vertical column of the image is converted into pulse trains of write voltages and sequentially applied to the reservoir's volatile memristors. (d) Parity preprocessing with 1D preprocessing: as in 1D, each horizontal row is converted into a pulse train of write voltages. In addition, the parity operation --- an XOR between adjacent rows --- produces an image with the outline of the digit. This image is also converted into pulse trains of write voltages and applied to separate memristors.}}
\label{fig_RC_preprocessing}
\vskip -2ex
\end{figure*}

Before entering the reservoir, each input image is converted into a sequence of spatio-temporal write pulses. These pulses are applied sequentially to the volatile memristors comprising the reservoir. We evaluate various preprocessing methods using the MNIST dataset, which includes 60,000 training and 10,000 test images. Each image comprises 28$\times$28 pixels with greyscale values from 0 to 255. These are binarized by labeling pixels above 25 as '1' (white) and others as '0' (black) to retain digit shape. In the reservoir, '1' translates to a write voltage pulse; '0' means no pulse \cite{rc_with_dynamic_memristors}.


All preprocessing methods generate pulse trains. When the write voltage pulse is applied (pixel value $'1'$), the internal state variable of the memristor is updated and during the remaining time (pixel value $'0'$), the memristor state decays. 
The number of pulse trains determines the number of volatile memristors in the reservoir. By employing different preprocessing techniques, the reservoir can capture distinct features of the input images, ultimately influencing the performance of the system.
This section details the different preprocessing approaches, some of which are introduced in \cite{rc_with_dynamic_memristors}, each of which translates pixels into memristor write pulses in a unique way. The reservoir size depends on the image dimensions, with the number of volatile memristors determined by the preprocessing method. Table~\ref{rc_size} summarizes the preprocessing methods discussed and their corresponding reservoir size.

\subsection{One Dimensional Data Input (1D)}

In this method, each \textit{horizontal} row of pixels in the image is converted into a pulse train of write voltages. Each pulse train is sequentially injected into a volatile memristor (see Fig.~\ref{fig_RC_preprocessing}(a)). If the input image is of size $n\times m$, where $n$ is the number of rows and $m$ is the number of columns, the reservoir will have $n$ volatile memristors, and each pulse train has $m$ pulses. 
In the case of MNIST, since the images have 28 rows, the reservoir has 28 volatile memristors.
Each volatile memristor accumulates the write pulses corresponding to $'1'$s in a given row; its volatile nature naturally decays between pulses, capturing not just the sum of $'1'$s but also their temporal distribution. The decay rate and the frequency of these input pulses must be calibrated to ensure an accurate row representation. However, using 1D input alone can limit classification performance because a single memristor may be overloaded by the entire row, reducing its ability to distinguish nuanced spatial features.

\subsection{\REV{Sectioning Input Rows}}
In input sectioning, each row and/or column of the image is divided into multiple smaller segments, as shown in Fig.~\ref{fig_RC_preprocessing}(b). Each segment (section) of pixels is converted into a shorter pulse train and sent to a separate volatile memristor, improving accuracy by reducing the amount of data each memristor must represent. For an input image of size $n \times m$ divided into $k$ sections, the reservoir size is $n\times k$ for 1D processing and $(n+m) \times k$ for 2D (Table \ref{rc_size}). For example, a $28\times28$ MNIST image with four sections per row in 1D mode requires $28 \times 4=112$ memristors, while 2D mode with six sections per row and column requires 
$(28+28) \times 6=336$ memristors. By reducing each memristor’s temporal load, sectioning significantly enhances the reservoir’s ability to capture intricate spatial features and thus improves classification accuracy.

\subsection{\REV{Two-Dimensional Data Input (2D)}}

In the 2D approach, both horizontal rows and vertical columns of the image are converted into pulse trains of write voltage, as shown in Fig.~\ref{fig_RC_preprocessing}(c). As in 1D data input, each pulse train is applied to a single volatile memristor.  For a 2D input of size $n \times m$, the reservoir requires $n+m$ memristors. For MNIST $(28\times28)$, this results in a reservoir size of $56$.
Because the image is applied from two orthogonal directions, the reservoir gains a richer spatial representation, boosting accuracy. The trade-off is a larger reservoir (double that of 1D). Furthermore, 2D preprocessing alone, like 1D, can be insufficient for classification if each memristor stores too many pulses. Hence, combining it with input sectioning ensures that each memristor receives a more manageable portion of the data. 

\section{Proposed Preprocessing Method - Parity}

The two-dimensional properties of the images are important but are not fully exploited without further preprocessing.
The proposed parity preprocessing method can enhance the network's accuracy by highlighting various features appearing in consecutive rows.  Specifically, it performs an XOR operation on the $i$-th and $(i+1)$-th rows, generating additional "parity rows". Because parity complements existing methods, it is used together with the 1D or 2D input -- as well as the input section -- to ensure a more faithful representation of the input data (see Fig.~\ref{fig_RC_preprocessing}(d)).
For instance, in 1D $+$ parity, the original rows are converted to write-voltage pulse trains, and the XORed rows are injected into separate memristors. 
With parity enabled, the 1D reservoir requires $2n-1$ volatile memristors, and the 2D case requires $(n+m+n-1)$. 
For MNIST $(28 \times 28)$, using four sections in 1D $+$ parity results in $(28 + 27)\times 4 = 220$ memristors, while 2D $+$ parity with six sections requires $(28 + 28 + 27) \times 6 = 498$ memristors. This method proves effective because the parity rows provide a sparse outline of the digits, helping to discern edges more clearly and thus improving classification performance. More sophisticated cross-row and column operations could be explored in future work, but here we employ the parity method because of its simplicity.

\section{Evaluation}

Before being converted into write pulses, each MNIST image is binarized and then transformed according to the chosen preprocessing method. For 1D, the image remains in its original $28\times28$ form. In 2D, the image is rotated by $90\degree$ and appended below the original, producing a $56 \times 28$ matrix. Similarly, in parity preprocessing, the XOR operation is performed on adjacent rows, and the resulting matrix is appended below the original image, extending the matrix by 27 rows. 
Finally, the rows are divided into sections, creating the pulse trains to be applied to the volatile memristors. 

Once the image has been converted into pulse trains, each pulse is multiplied by a write voltage ($V_{write} = 1.5V$, $t_{pulse} = 1 ns$) before being sent to the memristors. 
The pulse trains are then used to update the internal state variable $w$ of the memristor~\cite{model_of_Wox_memristor} according to Equations~(\ref{w_update}), (\ref{R(w)}), and ~(\ref{w_decay}), where $\alpha = 10^{-8} A$, $\beta = 0.5V^{-1}$, $\gamma = 10^{-5} A$, $\delta = 4V^{-1}$, $\lambda = 10^{3}s^{-1}$, $\eta=8V^{-1}$, $w_{Max} = 1$, $w_{Min} = 0.1$, and $\tau = 5 ns$. 
Each write pulse ('1') increases the memristor’s internal state, while the state gradually decays in the absence of pulses ('0'). After writing the entire image to the reservoir memristors, a single read pulse ($V_{read} = 0.6V$, $t_{pulse} =1 ns$) is sent simultaneously through each memristor of the reservoir, and the corresponding currents are obtained from Equation~(\ref{IV}). 

These read currents are \REV{first quantized to 6-bits, then rescaled using min-max scaling}, and multiplied by the readout layers' weight matrix. \REV{Quantization mimics an analog-to-digital converter (ADC) that may be used in hardware. It also helps assess the realistic accuracy that can be achieved in the presence of memristor variability.}
The readout layer was implemented as a logistic regression classifier, applying a sigmoid activation element-wise to the result of the matrix-vector product between the reservoir states and the weight matrix of dimensions $[N\times10]$, where N is the number of reservoir states, and 10 is the MNIST digit classes. This matrix is randomly initialized. 
During training, the predicted output for each input sample was computed and compared to a one-hot encoded label. The prediction error, equivalent to the gradient of the binary cross-entropy loss, was used to update the weights via stochastic gradient descent (SGD) using the outer product of the input vector and the error. This process was repeated for $500$ epochs with a learning rate of $0.02$. 

Only the readout-layer weight matrix is updated during training, using standard forward pass, error calculation, and gradient-based backpropagation. Accuracy is then measured on the 10,000-image test set. The network was implemented and trained in MATLAB. \REV{Memristor resolution was analyzed by quantizing the currents from 1-bit to 7-bits. Accuracy remains acceptable down to 4-bit quantization of memristor currents, but drops sharply below that.} 
In the following, we evaluate RC using accuracy, throughput, energy efficiency, and area, highlighting trade-offs among these metrics.
\subsection{Accuracy}

\begin{figure}[!t]
    \centerline{\includegraphics[scale=0.42]{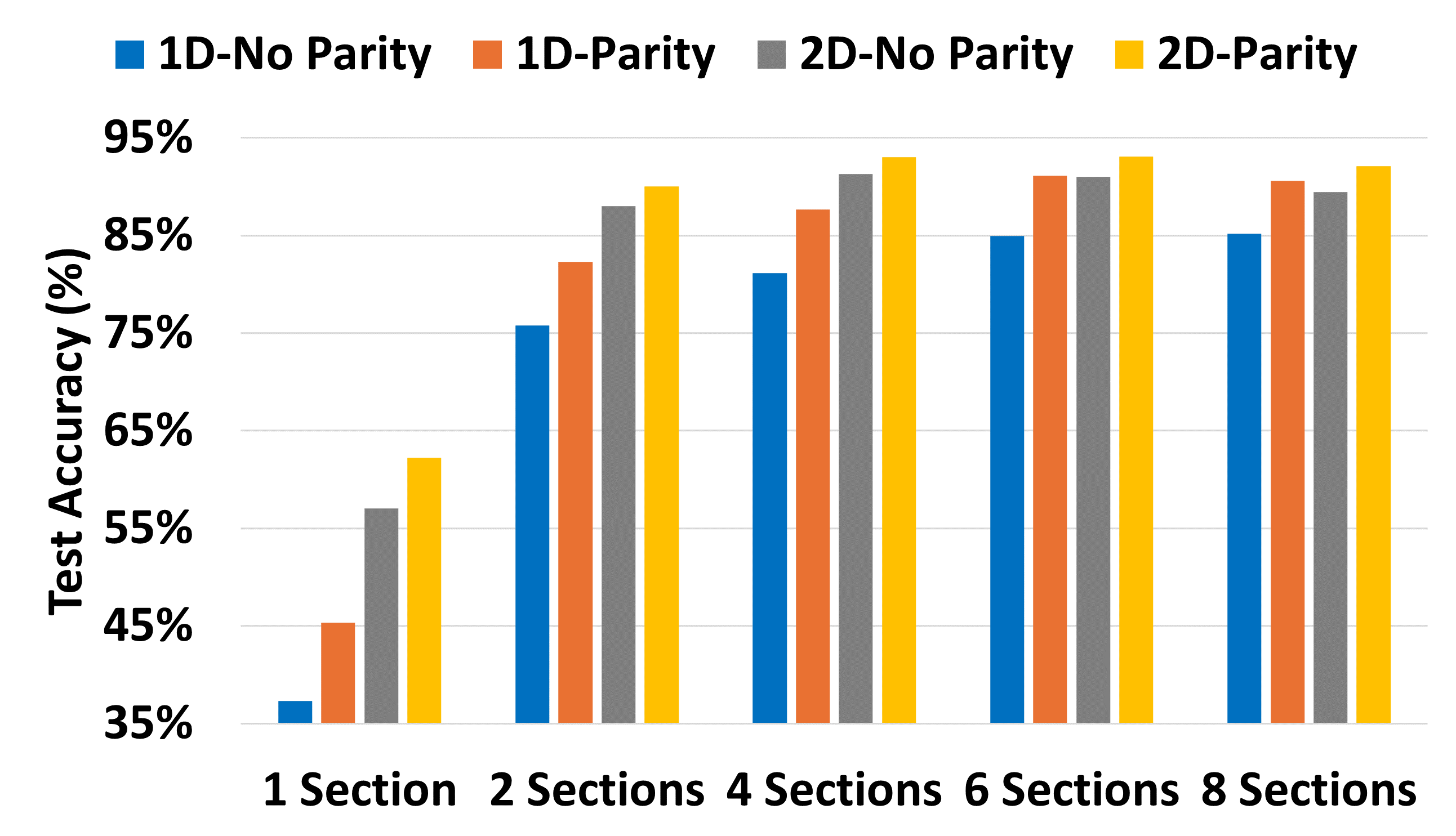}}
    \caption{Test accuracy across various preprocessing methods.}
    \label{accuracy}
\end{figure}

Fig.~\ref{accuracy} summarizes the test accuracy across preprocessing methods.
\REV{For the highest accuracy configuration (over 4 Sections),} 
2D improves the accuracy, on average, by 7\% (3\%) compared to 1D without (with) parity. Input sectioning proves essential in boosting accuracy, as each memristor receives a smaller, more manageable portion of the data. Without sectioning, the accuracy remains below 75\%. In 1D mode, adding more sections improves accuracy, but not sufficiently to match 2D or parity-enhanced configurations, whereas in 2D mode, excessive sectioning can over-fragment the representation and slightly reduce accuracy. Parity preprocessing yields a substantial accuracy gain --- approximately 6\% for 1D and 2\% for 2D --- by highlighting edges and offering richer input representation. 
\REV{As compared to \cite{rc_with_dynamic_memristors}, we achieve 1-2\% improvement in accuracy with fewer devices due to the parity method. While the accuracy is slightly lower than \cite{2mem_1cap_rc}, our method offers the advantage of not requiring large capacitors that limit the scalability of memristive systems.} 
\subsection{Throughput, Energy Efficiency, and Area}
\begin{figure*}[!t]
    \centerline{\includegraphics[scale=0.4]{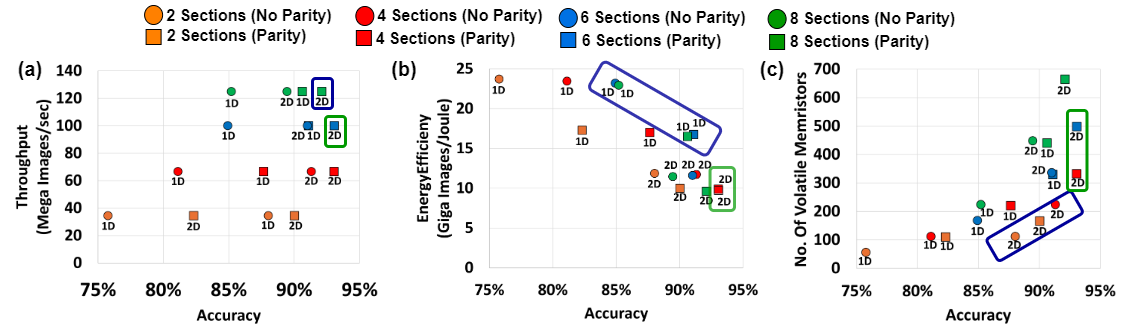}}
    \caption{
    Trade-offs between accuracy and (a) throughput, (b) energy efficiency, and (c) area. Methods using a single section with accuracy below 75\% are omitted for clarity. In (a) and (b), higher Y axis values are better; in (c), lower Y axis values are better. Green boxes indicate the most relevant, accurate configurations, while blue boxes highlight configurations offering a good balance between accuracy and the corresponding metric.
    }
    \label{trade-offs}
    \vskip -3.3ex
    \end{figure*}
\flushbottom
Here, throughput is determined as the number of images processed per second. The throughput grows almost linearly with the number of sections; it is independent of the number of dimensions and the usage of parity \REV{(see Fig. \ref{trade-offs} (a))}. 

The energy consumption of reservoir memristors is dominated by write operations (over $94\%$), as each memristor is read only once per image, and the read voltage ($0.6V$) is significantly lower than the write pulse voltage ($1.5V$). Energy consumed per image increases approximately linearly with the number of dimensions and is roughly independent of the number of sections. The parity method increases energy consumption sub-linearly, as parity-related rows contain fewer '1's than the original data, leading to fewer write operations \REV{(see Fig. \ref{trade-offs} (b))}. We define energy efficiency as the number of images processed per joule.

The relative area is approximated by the number of RC memristors. Each memristor \REV{also} includes the area of its input line (pulse train) and the output lines to the perceptron readout. The area grows roughly linearly with number of dimensions, use of parity, and number of sections \REV{(see Fig. \ref{trade-offs} (c))}.

Figure~\ref{trade-offs} illustrates the trade-offs between accuracy, throughput, energy efficiency, and area across different preprocessing methods. The results indicate that achieving the highest accuracy typically requires combining several complex preprocessing techniques, which, in turn, reduce throughput and energy efficiency, and increase area (marked green). In contrast, a modest reduction in accuracy can produce higher throughput and energy efficiency, and decrease area (marked blue). Among the methods evaluated, using parity generally achieves the highest accuracy, but performs less favorably in terms of throughput, energy efficiency, and area.

\section{Conclusion}

In this paper, we evaluated several preprocessing methods for memristive reservoir computing, using the MNIST dataset as a benchmark. We introduced a new parity-based preprocessing technique that improves recognition accuracy by enriching the reservoir's representation with additional spatial features. The use of novel volatile memristors to capture the spatio-temporal structure of the input data, significantly reduces hardware overhead compared to fully connected neural networks. These results highlight the potential of combining parity-based preprocessing with sections in memristive reservoir computing to enable accurate, efficient, high-performance neuromorphic systems. Future research may explore task-specific or adaptive preprocessing strategies to further enhance system performance across diverse applications.

\section*{Acknowledgment}

This research has been partially funded by the European Union's Horizon 2020 Research And Innovation Programme FET-Open NEU-Chip under grant agreement No. 964877 and funded by the European Union (ERC, Real-Database-PIM, 101157452). Views and opinions expressed are however those of the author(s) only and do not necessarily reflect those of the European Union or the European Research Council Executive Agency. Neither the European Union nor the granting authority can be held responsible for them.

\vspace{12pt}

\begin{thebibliography}{00}

\bibitem{next_gen_rc} D. J. Gauthier, E. Bollt, A. Griffith, and W. A. S. Barbosa, “Next generation reservoir computing,” \textit{Nat. Commun.}, Vol. 12, No. 1, p. 5564, Dec. 2021.

\bibitem{wind_forecasting_using_rc} Z. Tian, H. Li, and F. Li, “A combination forecasting model of wind speed based on decomposition,” \textit{Energy Rep.}, Vol. 7, pp. 1217–1233, Nov. 2021.

\bibitem{stock_market_prediction_using_rc} W.-J. Wang, Y. Tang, J. Xiong, and Y.-C. Zhang, “Stock market index prediction based on reservoir computing models,” \textit{Expert Syst. Appl.}, Vol. 178, pp. 115022, Sept. 2021.

\bibitem{img_recog_using_rc} Z. Tong and G. Tanaka, "Reservoir computing with untrained convolutional neural networks for image recognition," \textit{Proc. Int. Conf. Pattern Recognit. (ICPR)}, pp. 1289-1294, Beijing, China, Aug. 2018.

\bibitem{info_processing_in_single_dynamical_node} L. Appeltant \textit{et al.}, “Information processing using a single dynamical node as complex system,” \textit{Nat. Commun.}, Vol. 2, No. 1, pp. 1–6, Sept. 2011.

\bibitem{chaotic_time_series_using_rc} S. Shahi, F. H. Fenton, and E. M. Cherry, “Prediction of chaotic time series using recurrent neural networks and reservoir computing techniques: A comparative study,” \textit{Mach. Learn. Appl.}, Vol. 8, pp. 100300, June 2022.

\bibitem{physical_rc_with_emerging_electronics_review} X. Liang \textit{et al.},, “Physical reservoir computing with emerging electronics,” \textit{Nat. Electron.}, Vol. 7, pp. 193-206, Mar. 2024.

\bibitem{delay_based_rc_using_analog_ckts} M. C. Soriano \textit{et al.}, “Delay-based reservoir computing: Noise effects in a combined analog and digital implementation,” \textit{IEEE Trans. Neural Netw. Learn. Syst.}, Vol. 26, No. 2, pp. 388–393, Feb. 2015.

\bibitem{lsm_rc_using_fpga} Q. Wang, Y. Li, and P. Li, “Liquid state machine based pattern recognition on FPGA with firing-activity dependent power gating and approximate computing,” \textit{Proc. IEEE Int. Symp. Circuits Syst. (ISCAS)}, pp. 361-364, Montreal, QC, Canada, May 2016.

\bibitem{rc_with_dynamic_memristors} C. Du \textit{et al.}, “Reservoir computing using dynamic memristors for temporal information processing,” \textit{Nat. Commun.}, Vol. 8, No. 1, p. 2204, Dec. 2017.

\bibitem{photonic_rc_review} G. Van Der Sande, D. Brunner, and M. C. Soriano, “Advances in photonic reservoir computing,” \textit{Nanophotonics}, Vol. 6, No. 3., pp. 561–576, May 2017.

\bibitem{spintronic_rc} A. J. Edwards \textit{et al.}, “Passive frustrated nanomagnet reservoir computing,” \textit{Commun. Phys.}, Vol. 6, No. 1, p. 215, Dec. 2023.

\bibitem{echo_state_graph_neural_network} S. Wang \textit{et al.}, “Echo state graph neural networks with analogue random resistive memory arrays,” \textit{Nat. Mach. Intell.}, Vol. 5, No. 2, pp. 104–113, Feb. 2023.

\bibitem{snn_memristive_rc} R. A. John \textit{et al.}, “Reconfigurable halide perovskite nanocrystal memristors for neuromorphic computing,” \textit{Nat. Commun.}, Vol. 13, No. 1, p. 2074, Dec. 2022.

\bibitem{rc_book_theory_physical_implementations_applications} K. Nakajima and I. Fischer, Natural Computing Series Reservoir Computing Theory, Physical Implementations, and Applications, \textit{Springer Singapore}, Aug. 2021. 

\bibitem{jaeger_echo_state_paper} H. Jaeger, "The "echo state" approach to analysing and training recurrent neural networks-with an erratum note'," \textit{Bonn, Germany: German National Research Center for Information Technology GMD Technical Report}, 148, Jan. 2001.  

\bibitem{volatile_and_non_volatile_memristive_devices_for_neuromorphic_computing} G. D. Zhou \textit{et al.}, "Volatile and nonvolatile memristive devices for neuromorphic computing," \textit{Adv. Electron. Mater.}, Vol. 8, No. 7, p. 2101127, Feb. 2022.

\bibitem{model_of_Wox_memristor} T. Chang, \textit{et al.}, "Synaptic behaviors and modeling of a metal oxide memristive device," \textit{Appl. Phys. A}, Vol. 102, pp. 857–863, Feb. 2011. 

\bibitem{novel_memristor_esn_MEM-ESN} J. Sun \textit{et al.}, "A novel memristors based echo state network model inspired by the brain’s uni-hemispheric slow-wave sleep characteristics," \textit{Cogn. Comput.}, Vol. 16, pp. 1470–1483, June 2024. 

\bibitem{hardware_esn_with_double_crossbar_memristors} A. M. Hassan, H. H. Li and Y. Chen, "Hardware implementation of echo state networks using memristor double crossbar arrays," \textit{ Proc. 2017 Int. Joint Conf. Neural Networks (IJCNN)}, pp. 2171-2177, Anchorage, AK, USA, May 2017. 

\bibitem{memristor_based_LSM_with_insitu_training_method} A. Henderson \textit{et al.}, "Memristor based liquid state machine with method for in-situ training," \textit{IEEE Trans. Nanotechnol.}, Vol. 23, pp. 376-385, Mar. 2024.

\bibitem{non_masking_based_rc_with_single_dynamic_memristor} X. Wu \textit{et al.}, "Nonmasking-based reservoir computing with a single dynamic memristor for image recognition," \textit{Nonlinear Dyn.}, Vol. 12, pp. 6663–6678, Mar. 2024. 


\bibitem{2mem_1cap_rc} \REV{S. K. Shim \textit{et al.}, “2Memristor-1Capacitor integrated temporal kernel for high-dimensional data mapping,” Small, Vol. 20, No. 25, p. 2306585, June 2024.}
\end{thebibliography}
\end{document}